\documentclass[mlmain]{jmlr}
\jmlrproceedings{Preprint}{Preprint}

\usepackage{longtable}

\usepackage{booktabs}
\usepackage[load-configurations=version-1]{siunitx} 

\theorembodyfont{\upshape}
\theoremheaderfont{\scshape}
\theorempostheader{:}
\theoremsep{\newline}

\jmlrpages{}
\firstpageno{1}
\editors{}
\jmlryear{2026}
\jmlrworkshop{}

\title[Concurrent Temporal Variables in Language Models]{Factorized Representations of Concurrent Temporal Variables in Pretrained Language Models}

  \author{\Name{Xinhe Zhang} \Email{xinhezhang@g.harvard.edu} \\
  \Name{Arnau Marin-Llobet} \Email{amarinllobet@seas.harvard.edu}\\
  \addr School of Engineering and Applied Sciences, Harvard University, Cambridge, USA.}

\begin{document}

\maketitle

\begin{abstract}
Language models often need to maintain multiple contextual variables that evolve independently within a document. Prior work has identified structured representations of individual temporal or contextual quantities, but their joint organization remains less understood. We study this question using controlled documents with independently phased language-switch and topic-change timers. Across ten pretrained checkpoints, both timers are linearly decodable from the same hidden states. Removing a timer-associated centroid plane substantially reduces access to its associated variable while having little effect on the concurrent timer, although the planes are nonorthogonal and contain cross-timer information. Separate timer contributions also explain most reproducible variation among joint-state centroids. These signatures persist from 3 to 51 timer values and across Qwen3 checkpoints from $0.6$B to $14$B. Schedule position affects continuation probabilities, but the tested centroid displacement produces no consistent timer-specific output effect. Together, these results support approximate factorization based on selective accessibility and approximately additive joint-state geometry in overlapping subspaces.
\end{abstract}

\section{Introduction}
\label{sec:intro}

A language model processing a document may need to maintain several properties of the context that evolve independently. In dialogue, the speaker may change while the topic remains fixed; in multilingual text, the language may switch independently of the subject matter. At each position, the model must represent the current values of these variables while updating them according to different histories. We study how such concurrent, context-defined variables are jointly organized in pretrained hidden representations.

Prior work has identified spatial and temporal coordinates \citep{gurnee2024space}, contextual binding vectors \citep{feng2024binding}, and a countdown subcircuit \citep{dunefsky2026shared}. These studies establish structured representations of individual temporal or contextual quantities. Much less is known about multiple variables that evolve independently within the same document. Simultaneous linear decodability does not determine whether two variables rely on shared or variable-specific directions, and subspace overlap does not determine whether the information in that overlap is required for decoding.

We use controlled documents with independently phased language switches and topic changes. Each variable is represented by the number of sentences since its most recent event. The two schedules produce balanced combinations of timer values while preserving ordinary sentence content. For each timer, we estimate a low-dimensional subspace from its class centroids. We then remove this subspace, retrain the linear decoder, and measure how much timer information remains accessible. Comparing removal of a timer's own plane with removal of the concurrent timer's plane provides a direct measure of selectivity.

Across ten pretrained checkpoints, removing each timer's centroid plane eliminates substantially more of its associated decoding signal than removing the concurrent timer's plane. The planes are nonorthogonal and contain information about both timers, while separate timer contributions explain most joint-centroid variation that generalizes across document splits. These results characterize \emph{approximate factorization} in pretrained representations through selective accessibility and approximately additive joint-state geometry, with overlapping factor-associated subspaces.

We further test generalization across timer horizon and model size. Across 3 to 51 timer values, decoding remains high, effective centroid rank grows slowly, and self-ablation remains much larger than cross-ablation, although cross-ablation increases with horizon. Within Qwen3 from $0.6$B to $14$B, the same signatures persist without a monotonic parameter trend. A recency-matched continuation assay shows that schedule position affects model outputs. Centroid displacements reliably alter decoded timer states, but the tested language-timer displacement from value 1 to value 2 produces no consistent timer-specific change in continuation probabilities across models.

\section{Experimental Framework}

\subsection{Concurrent event-recency variables}

We construct controlled documents with independently phased language switches and topic changes. English and German sentences are drawn from a hand-written bank of parallel propositions on two topics, house and field. No timer labels, event markers, prefixes, or separators are added. Language switches occur every three sentences and topic changes every four, with independent document-level offsets.

The language timer $y_{\mathrm{lang}}(t)\in\{0,1,2\}$ records sentences since the most recent switch. The topic analysis retains phases $0$--$2$ from its four-phase schedule. Independent phases and coprime periods yield balanced, independent analyzed labels (Figure~\ref{fig:selectivity}A).

The primary corpus contains 100 documents of 48 sentences. Proposition identity is sampled independently within the active topic. Additional analyses use matched stochastic schedules and larger corpora (Appendix~\ref{app:details}).

\subsection{Hidden-state representations and linear readout}

We evaluate ten pretrained checkpoints from eight model families, including base and instruction-tuned models: Llama-3.2-1B, Mistral-7B-v0.3, Qwen3-1.7B, Qwen3-1.7B-Base, Qwen3-4B, pythia-1.4b, gemma-2-2b, SmolLM2-1.7B, Phi-3-mini, and OLMo-2-1B. Model parameters remain fixed. For each sentence, we extract the sentence-final residual-stream state $h_t$ from layer $\lfloor L/2\rfloor$, where $L$ is the number of transformer blocks. Table~\ref{tab:model_checkpoints} lists the exact checkpoints and analyzed layer indices.

We subtract each document's mean state and residualize each hidden dimension against $\{1,t,t^2\}$ in sentence index $t$. These nuisance-normalization steps use full-document and corpus-level statistics and precede cross-fitting. Documents are then divided into disjoint roles for fitting a 100-component PCA representation and timer planes, training ridge decoders, and evaluation. Within each split, PCA is fitted only on the plane-estimation documents and applied to the other document sets.

\subsection{Timer-associated subspaces}

For timer value $v\in\{0,1,2\}$, let $\mu_v$ denote the plane-estimation-fold mean for examples with that value. After subtracting their common mean, the three class centroids span at most two dimensions. We denote this rank-2 \emph{centroid plane} by $S_e$ for event timer $e$.

Projection outside the centroid plane eliminates between-class mean differences on the plane-estimation fold. Cross-fitting evaluates the transfer of this geometry to disjoint documents, and post-ablation decoding measures the timer information that remains linearly accessible after decoder retraining. Plane stability is evaluated across disjoint document subsets and against shuffled labels. All reported planes are mapped back to the model's state space before geometric comparison or intervention.

To measure how much linearly accessible information remains outside a subspace $S$, we project
\[
h'_t=(I-P_S)h_t,
\]
where $P_S$ is the orthogonal projector onto $S$, and retrain the decoder. Retraining measures information that remains linearly accessible after projection. One document subset fits the plane, a second trains intact and post-ablation decoders, and a third is evaluated; roles rotate across splits. Ablation is reported as the fraction of above-chance decoding signal removed.

\subsection{Selective ablation, information overlap, and joint additivity}
\label{sec:measures}

\paragraph{Selective ablation.}
For timers $A$ and $B$, \emph{self-ablation} removes $S_A$ before decoding $A$; \emph{cross-ablation} removes $S_A$ before decoding $B$. Decoders are retrained in both conditions.

\paragraph{Information overlap.}
Restricted decoding trains a decoder using only the coordinates within each plane. Principal angles measure geometric alignment between planes. Shuffled-label planes provide controls for both quantities.

\paragraph{Joint additivity.}
For the nine analyzed combinations $(a,b)$ of the two timer values, we fit
\[
\mu(a,b)=m+u_a+v_b+\epsilon_{ab},
\]
where $m$ is a common mean, $u_a$ and $v_b$ are timer contributions, and $\epsilon_{ab}$ is their interaction. Interaction is the fraction of joint-centroid variance attributable to $\epsilon_{ab}$. We calibrate it with finite-sample surrogates whose population centroids are exactly additive and with shuffled timer labels. Cross-document fits compare additive predictions with a saturated nine-centroid model.

\subsection{Horizon, parameter-range, and behavioral extensions}

For $H\in\{2,5,10,20,50\}$, schedules with periods $H+1$ and $H+2$ yield analyzed timer values $0,\ldots,H$. Each horizon contains 200 documents generated from an expanded parallel proposition bank. Ridge regression predicts timer value, with performance evaluated by held-out Spearman correlation, and the centroid rank is selected to explain $90\%$ of centroid variance on the plane-estimation fold. Self-, cross-, and random-subspace ablations use the same selected rank under three-way cross-fitting.

We also evaluate five Qwen3 sizes from $0.6$B to $14$B under a common quantization and analysis protocol. The behavioral assay is evaluated on the same ten pretrained checkpoints as the primary analysis and scores the summed log-probability difference between parallel English and German candidate continuations under period-3 and period-4 schedules at matched language recency. Language-timer centroid displacement is applied at the sentence-final token and compared with matched-norm concurrent-timer and isotropic displacements (Appendix~\ref{app:extensions}).

\section{Concurrent Temporal Variables Exhibit Selective Ablation Effects}
\label{selective_ablation}

\subsection{Both temporal variables are simultaneously represented}

We first characterize the linear accessibility of both timers from the same hidden states. We also repeat the analysis after excluding current-event sentences to distinguish elapsed-state information from direct detection of a switch or topic change.

Across all ten models, linear decoders predict both timers above the $33.3\%$ chance level. Balanced accuracy is $84$--$96\%$ for language and $40$--$53\%$ for topic; the topic signal is weaker but consistently above chance.

After excluding current-event states, value-1 versus value-2 accuracy is $89.3\%$ for language and $59.8\%$ for topic, against $50\%$ chance. Both timers remain decodable among past-event states.

\subsection{Timer-associated planes selectively capture accessible timer information}

For each timer, we compare intact decoding with decoding after removing either its own plane or the concurrent timer's plane. Decoders are retrained after projection, and plane estimation, decoder training, and evaluation use disjoint document roles, so the comparison measures how much linearly accessible information remains outside each independently estimated plane.

\begin{figure}[!htbp]
\centering
\setlength{\abovecaptionskip}{2pt}
\setlength{\belowcaptionskip}{0pt}
\includegraphics[width=\linewidth]{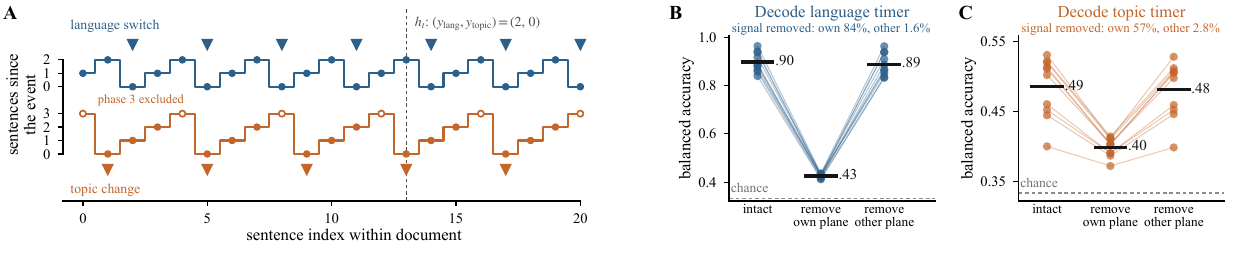}
\vspace{-20pt}
\caption{\textbf{Selective ablation of concurrent timer representations.}
\textbf{A}, Independently phased language and topic schedules.
\textbf{B--C}, Cross-fitted decoding after subspace removal and decoder
retraining. Points represent models; bars indicate means.}
\label{fig:selectivity}
\end{figure}

Cross-fitted self-ablation of the language plane reduces accuracy from $0.897$ to $0.426$, removing $84\%$ of above-chance signal; topic-plane ablation leaves accuracy at $0.888$ and removes $1.6\%$. For the topic timer, self-ablation reduces accuracy from $0.486$ to $0.399$ and removes $57\%$, whereas language-plane ablation leaves accuracy at $0.481$ and removes $2.8\%$ (Figure~\ref{fig:selectivity}B--C). The self--cross separation occurs in each model despite variation in intact decoding accuracy.

\subsection{Selective ablation under additional conditions}

In the controlled-overlap calibration, cross-ablation increases as timer-specific signal is transferred into the concurrent plane. Similar self--cross separation is observed with phase-matched document splits, aperiodic schedules, alternative plane estimators and PCA dimensions, and DailyDialog \citep{li2017dailydialog}. Mean per-cell cross/self ratios are $3.40\%$ for aperiodic schedules and $3.51\%$ for matched periodic schedules; on DailyDialog, mean cross-ablation damage is $12.6\%$ of self-ablation damage. History-only analyses show the same separation. Appendices~\ref{app:selective} and~\ref{app:extensions} provide the corresponding results.

\section{Concurrent Temporal Variables Form an Approximately Factorized Joint Representation}

To characterize the joint representation beyond selective ablation, we examine how the timer planes overlap and whether the nine joint timer states combine approximately additively. We measure overlap using restricted decoding and principal angles, and additivity from the joint-centroid geometry.

\subsection{Timer-associated planes are not information-disjoint}

\begin{figure}[!htbp]
\centering
\setlength{\abovecaptionskip}{2pt}
\setlength{\belowcaptionskip}{0pt}
\includegraphics[width=\linewidth]{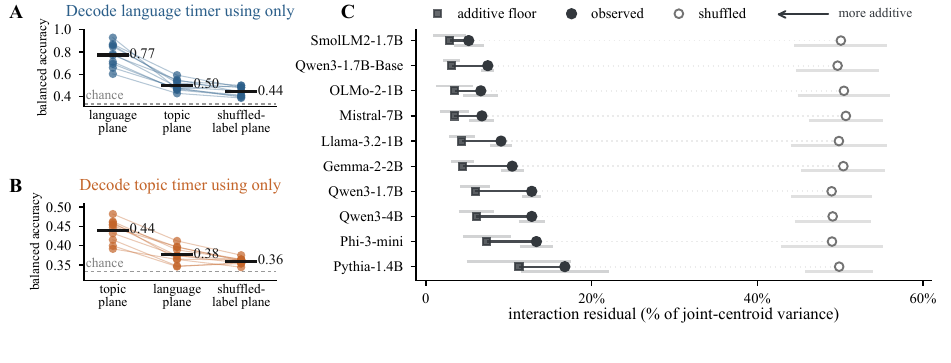}
\vspace{-20pt}
\caption{\textbf{Information overlap and approximately additive joint geometry.}
\textbf{A--B}, Decoding restricted to timer-associated planes.
\textbf{C}, Joint-centroid interaction residuals across models.
Pale bars indicate the mean $\pm$ bootstrap standard error.}
\label{fig:factorization}
\end{figure}

Restricted decoding from a rank-2 plane shows that each timer is most informative in its own plane but is also decodable from the concurrent plane. Language decoding is $0.440$ above chance from its own plane and $0.166$ from the topic plane; topic decoding is $0.107$ above chance from its own plane and $0.043$ from the language plane. In both directions, concurrent-plane decoding exceeds the shuffled-label control (Figure~\ref{fig:factorization}A--B). Restricted decoding therefore shows that each timer plane contains information about the concurrent timer.

The smaller principal angle between timer planes averages $53.9^\circ$, compared with $79.7^\circ$ for unrelated rank-2 subspaces in the same 100-dimensional representation. The planes are therefore nonorthogonal as well as information-overlapping. Together with the ablation result, this shows that representational overlap and selective accessibility can coexist.

\subsection{Shared geometry contains common and event-specific structure}

Whitening within-class scatter reduces both plane alignment and cross-timer decoding, indicating that shared covariance contributes to the observed overlap. The most closely aligned direction carries approximately $47\%$ of the full language-plane decoding signal. Removing this direction increases the angle between the remaining plane directions from $54.8^\circ$ to $78.3^\circ$, revealing more nearly orthogonal residual structure (Appendix~\ref{app:joint}).

Across six event types and four models, a rank-2 plane fitted after pooling across event types recovers $78.5\%$ of the decoding signal from event-specific planes. Event-specific residual directions outperform matched random subspaces in all 24 model--event combinations. Together, these analyses identify shared event-recency structure alongside event-specific geometry (Appendix~\ref{app:extensions}).

\subsection{The joint timer geometry is approximately additive}

We next test whether changing one timer produces a similar centroid displacement at each value of the other timer. Departures from this additive structure are quantified by the interaction residual defined in Section~2.4.

Interaction accounts for $5.2$--$16.8\%$ of joint-centroid variance, with a mean of $10.1\%$. Exactly additive finite-sample surrogates yield $5.2\%$ mean interaction, and shuffled labels yield approximately $50\%$ (Figure~\ref{fig:factorization}C). The joint geometry therefore contains measurable non-additivity, while additive timer contributions account for most centroid variation.

Across document halves, the additive model explains $81.0\%$ of held-out joint-centroid variance, compared with $78.2\%$ for a saturated nine-centroid model. The saturated model exceeds the additive model in three of ten checkpoints, by at most $0.007$ explained variance. The predominantly additive structure thus generalizes across document splits.

The two timers differ in geometric reliability. Language-timer displacements are more reproducible across document splits and more stable across concurrent-timer values than topic-timer displacements, consistent with the stronger language-timer decoding signal (Appendix~\ref{app:joint}).

\subsection{Timer-plane displacements selectively alter decoded timer states}

The additive model associates a displacement with each change in timer value. We apply these displacements to held-out states to test whether they alter the corresponding decoded variable while preserving the concurrent timer.

For a held-out state with timer value $i$, we add the within-plane displacement from centroid $\mu_i$ toward $\mu_j$. A full displacement produces the target decoded value in $94.2\%$ of cases, compared with $6.6\%$ for an equal-norm isotropic displacement. At half displacement, the decoded target timer changes in $47.6\%$ of cases, while the decoded concurrent timer remains unchanged in $93.2\%$. All six ordered source-to-target transitions contribute to these aggregates. These results concern linear readouts of the intervened hidden state. Section~\ref{sec:generalization_behavior} tests whether one such intervention, the language-timer displacement from value 1 to value 2, produces a corresponding change in continuation probabilities.

\section{Generalization and Behavioral Relevance}
\label{sec:generalization_behavior}

We next examine whether the identified representational organization persists across longer timer horizons and model sizes, and whether it is reflected in model outputs.

\subsection{Extension to longer timer horizons}

With three timer values, the centered class centroids necessarily span at most two dimensions. We therefore repeat the analysis with 3, 6, 11, 21, and 51 timer values, selecting the smallest rank that explains $90\%$ of centroid variance on the plane-estimation fold.

\begin{table}[!htbp]
\centering
\small
\begin{tabular}{rrrrrrrr}
\toprule
$H$ & Values & $\rho$ & Rank & Self & Cross & Random & Cross/self \\
\midrule
2  & 3  & 0.91 & 1.75 & 84.9\% & 0.15\% & 0.18\% & 0.18\% \\
5  & 6  & 0.89 & 2.50 & 75.3\% & 0.44\% & 0.31\% & 0.56\% \\
10 & 11 & 0.87 & 3.25 & 79.9\% & 2.05\% & 0.53\% & 2.58\% \\
20 & 21 & 0.87 & 3.48 & 79.8\% & 2.96\% & 0.69\% & 3.71\% \\
50 & 51 & 0.86 & 3.98 & 84.6\% & 4.86\% & 0.76\% & 5.71\% \\
\bottomrule
\end{tabular}
\caption{Cross-fitted horizon sweep across four models.
Rank is the mean selected dimension; Self, Cross, and Random
are percentage reductions in intact Spearman correlation.
See Appendix~\ref{app:horizons} for aggregation details.}
\label{tab:horizon_crossfit}
\end{table}

Across four models, decoding remains high as the number of timer
values increases, while selected centroid rank grows slowly
(Table~\ref{tab:horizon_crossfit}). Self-ablation remains much
larger than cross-ablation throughout the sweep. Cross-ablation
increases with horizon and exceeds matched-rank random ablation
at longer horizons, indicating gradually weaker selectivity.

\subsection{Persistence across model sizes}

To examine variation with model size within a single model family, we repeat the analysis across Qwen3 checkpoints at $0.6$B, $1.7$B, $4$B, $8$B, and $14$B while holding the corpus, layer-selection rule, preprocessing, and analysis protocol fixed. The five 4-bit checkpoints form the primary comparison, with bfloat16 runs at the three overlapping sizes as a precision control.

Using the full-corpus interaction estimator at each model size, interaction residuals remain between $12.1\%$ and $15.9\%$, well below shuffled values near $50\%$. Language-timer split-half plane angles remain between $17^\circ$ and $27^\circ$, compared with $84^\circ$--$86^\circ$ for shuffled labels (Table~\ref{tab:qwen_ladder}); decoding and selective ablation also persist at every size. Neither interaction residual nor language-plane split-half angle varies monotonically with parameter count; their Spearman correlations with parameter count are $+0.10$ and $+0.20$, respectively. The representational signatures therefore persist across model sizes without a monotonic scaling trend.

\subsection{Relation to continuation behavior}

We next examine the relation between timer representations and the model's continuation distribution. We first test whether continuation probabilities depend on schedule position by comparing period-3 and period-4 language-switch schedules at matched immediate language recency. Parallel English and German candidate sentences express the same proposition and topic, and a difference-in-differences contrast isolates the effect of schedule position beyond immediate recency. This contrast is positive in all ten pretrained models, ranging from $1.78$ to $4.14$ in candidate-sentence log odds, while a randomly initialized Qwen3-1.7B has the opposite sign (Table~\ref{tab:behavior_gate}). Continuation probabilities therefore depend on schedule position beyond immediate language recency.

We then test whether the identified timer displacement can selectively alter this behavior. The language-timer displacement from value 1 to value 2 is applied at the middle-layer sentence boundary, at strengths $0.5$ and $1.0$, and compared with matched-norm concurrent-timer and isotropic displacements. The target displacement does not produce a consistently larger continuation effect than the matched concurrent-timer displacement. At full strength, target-minus-concurrent effects range from $-0.033$ to $+0.029$, average $-0.0046$, and are positive in five of ten models (Table~\ref{tab:behavior_steer}); the same qualitative pattern holds at half strength.

Schedule position is reflected in the model's continuation distribution, whereas the tested centroid displacement changes decoded timer state without a consistent timer-specific continuation effect.

\section{Related Work}

\subsection{Factorized and structured representations}

Transformers trained on controlled stochastic processes can represent predictive belief-state geometry \citep{shai2024belief}, and models trained on independent synthetic factors can develop approximately orthogonal factor subspaces \citep{shai2026factored}. \citet{thil2026subspace} partition {JEPA} latents into designated progression and content subspaces and evaluate the partition through ablation. These studies examine factors specified by the training process or architecture. We analyze independently evolving contextual variables in fixed pretrained models and characterize their organization through selective accessibility and additive joint-state geometry, including nonorthogonal, information-overlapping subspaces.

\citet{fel2026structuring} model activations as sparse additive mixtures of low-dimensional concept manifolds and introduce block-sparse featurizers for discovering this structure without concept labels. Our analyses use experimentally specified variables to measure how their information is distributed across subspaces and how their joint states combine.

\subsection{Temporal state and representational intervention}

Language models encode spatial and temporal coordinates \citep{gurnee2024space}, distance to a target length \citep{dunefsky2026shared}, repeated-token count \citep{venkatesh2026counting}, and contextual entity bindings \citep{feng2024binding}. Elapsed wall-clock time can also be supplied through a trained conditioning channel \citep{siavoshian2026chronostate}. These studies principally address individual temporal quantities or variables introduced and later queried within a context. The present experiments examine two temporal variables that are repeatedly updated in the same hidden states.

Concurrent working-memory variables can exhibit representational overlap \citep{xiong2026incontextsuperpositionhumanlikeworking} and behavioral interference \citep{xiong2026incontextsuperpositionhumanlikeworking,wang2025unable}. Causal abstraction and interchange-intervention methods evaluate whether representational interventions alter a target while preserving other attributes \citep{huang2024ravel,wu2023interpretability}; control tasks calibrate probe capacity \citep{hewitt2019control}. Our analyses distinguish information contained within a timer plane, information remaining after its removal, and output changes under centroid displacement.

\section{Discussion and Conclusion}

The results support an operational notion of approximate factorization based on selective accessibility and approximately additive joint-state geometry \citep{park2024linear, fel2026structuring}. The timer subspaces are nonorthogonal and information-overlapping, and each contains information about both variables. Nevertheless, removing each timer plane disproportionately impairs decoding of its associated timer, while additive timer contributions explain most joint-centroid variation that generalizes across document splits. Together, restricted decoding, ablation, and joint-state geometry show that factorized organization can coexist with distributed and overlapping representations.

This organization persists as the number of timer values increases from 3 to 51 and across Qwen3 checkpoints from $0.6$B to $14$B. Effective centroid rank grows slowly with horizon, while cross-ablation increases but remains substantially weaker than self-ablation. Recency-matched schedule position systematically affects continuation probabilities, and centroid displacements reliably alter decoded timer states, but the tested value-1-to-value-2 intervention produces no consistent timer-specific change in model output. Thus, selective manipulation of decoded state does not necessarily translate into selective downstream control.

These results motivate further study of when factorized representational structure supports controllable computation. Extending this analysis to richer concurrent variables, more naturalistic contexts, and interventions distributed across layers or multiple tokens may clarify which representational structures contribute to downstream computation. Furthermore, bridging these activation-level timer geometries to underlying counting subcircuits \citep{dunefsky2026shared} and weight-sparse transformer mechanisms \citep{gao2025weight, marin2026individual} could reveal how these states are physically implemented. More broadly, relating factorization to intervention, causal influence, and task-dependent control could help distinguish variables that are accessible in hidden states from those that exert measurable influence on model behavior. 

\clearpage 

\section*{Acknowledgements}
Arnau Marin-Llobet is supported by Coefficient Giving and the RCC-Harvard Fellowship.

\bibliography{references}

\clearpage

\appendix

\section{Experimental Details}
\label{app:details}

\paragraph{Controlled corpus construction.}
The generator uses a hand-written bank of eight propositions for each of two topics, \emph{house} and \emph{field}, with parallel realizations in English, German, and Spanish. The primary corpus uses the 32 English and German sentence strings. Each sentence contains four whitespace-delimited words. Each document concatenates 48 sentences with single spaces, without added labels or markers.

In family A, language alternates between English and German every three sentences, while topic alternates every four sentences, with independent random document-level offsets. At each sentence, a proposition is sampled independently from the eight within the active topic. Documents share the proposition bank, and splits are document-disjoint. A matched family B uses stochastic language switches for additional schedule analyses. Each family contains 100 documents; primary analyses use family A.

Topic phase takes values $0, 1, 2$, and $3$. Primary three-class topic analyses retain phases 0–2, corresponding to 0, 1, or 2 sentences since the topic change.

\begin{table}[!htbp]
\centering
\small
\begin{tabular}{r p{0.30\textwidth} l l cc}
\toprule
Sent. & Text & Lang. & Topic & $y_{\mathrm{lang}}$ & $y_{\mathrm{topic}}$ \\
\midrule
0 & The lamp is bright.      & EN & house & 1 & 3$^\dagger$ \\
1 & The mountain is high.    & EN & field & 2 & 0 \\
2 & Der Fluss ist breit.     & DE & field & 0 & 1 \\
3 & Der Stein ist grau.      & DE & field & 1 & 2 \\
4 & Der Wald ist dunkel.     & DE & field & 2 & 3$^\dagger$ \\
5 & The chair is heavy.      & EN & house & 0 & 0 \\
6 & The lamp is bright.      & EN & house & 1 & 1 \\
7 & The window is open.      & EN & house & 2 & 2 \\
\bottomrule
\end{tabular}
\caption{
Excerpt from a family-A document with language and topic timer labels. \\$^\dagger$Topic-phase-3 rows are excluded from primary three-class topic analyses.}
\label{tab:controlled_example}
\end{table}

\paragraph{Representation extraction.}
Each full document is tokenized and processed in one forward pass, returning hidden states at every layer. For each sentence, we select the last nonzero-width token whose tokenizer offset ends at or before the sentence-final character boundary recorded by the generator. We analyze the residual-stream state at layer $\lfloor L/2\rfloor$, where $L$ is the number of transformer blocks. In the primary ten-checkpoint analysis, models run in bfloat16; extracted states are stored in float32 and converted to float64 for analysis.

\paragraph{Preprocessing.}
Before splitting documents for analysis, we subtract each document's mean sentence state and remove positional trends by regressing each hidden dimension on the corpus-wide design matrix $\{1,t,t^2\}$, where $t$ is sentence index. These steps use full-document and corpus-level statistics. Subsequent PCA, timer-plane estimation, decoder training, and evaluation follow the document-disjoint cross-fitting procedure. PCA is fitted on the designated fit documents, and the fitted basis is applied to decoder-training and evaluation documents.

\paragraph{Timer-associated planes and decoding.}
For each timer value $v\in\{0,1,2\}$, we compute the mean representation $\mu_v$ on the plane-estimation fold. After subtracting their common mean, the three centroids span at most two dimensions; their nonzero right singular vectors define the timer's centroid plane. Linear timer decoders are ridge classifiers fitted to one-hot targets with regularization $\lambda=10^{-2}n$, where $n$ is the number of decoder-training examples. We report balanced accuracy for classification.

To ablate a subspace $S$, we project states as $h'=(I-P_S)h$ and retrain the decoder on the projected states. Post-ablation accuracy measures timer information that remains linearly accessible outside $S$. For the primary cross-fitted ablation summary, we average intact and post-ablation accuracies across models and calculate the percentage of above-chance signal removed as
$
100\frac{\bar a_{\mathrm{intact}}-\bar a_{\mathrm{ablated}}}
{\bar a_{\mathrm{intact}}-a_{\mathrm{chance}}}.
$

\paragraph{Cross-fitted ablation.}
The 100 family-A documents are repeatedly partitioned into three disjoint subsets: one fits PCA and both timer planes, one trains intact and post-ablation decoders, and one evaluates them. Rotating these roles yields 36 splits per model and ablation direction. In-sample calibration estimates the plane and trains the decoder on the same subset, using the same number of plane-estimation documents as cross-fitting.

\paragraph{Additive interaction controls.}
To estimate finite-sample residuals under exact additivity, we combine each joint state's empirical residuals with its fitted additive mean, resample documents, and re-estimate the nine centroids. This surrogate has exactly additive population means and preserves empirical residual variability. For held-out prediction, we fit the additive model on half the documents and evaluate it against joint centroids from the other half, alongside mean-only and saturated nine-centroid models.

\paragraph{History-only decoding.}
History-only decoding retains target timer values 1 and 2, giving a binary task with $50\%$ chance accuracy. Timer planes and decoders are re-estimated on this subset.

\paragraph{Models.}

\begin{table}[!htbp]
\centering
\small
\begin{tabular}{l p{0.44\textwidth} rr}
\toprule
Paper name & Checkpoint & Layers & Analyzed \\
\midrule
Llama-3.2-1B & \texttt{meta-llama/Llama-3.2-1B} & 16 & 8 \\
Mistral-7B-v0.3 & \texttt{mistralai/Mistral-7B-v0.3} & 32 & 16 \\
Qwen3-1.7B & \texttt{Qwen/Qwen3-1.7B} & 28 & 14 \\
Qwen3-1.7B-Base & \texttt{Qwen/Qwen3-1.7B-Base} & 28 & 14 \\
Qwen3-4B & \texttt{Qwen/Qwen3-4B} & 36 & 18 \\
pythia-1.4b & \texttt{EleutherAI/pythia-1.4b} & 24 & 12 \\
gemma-2-2b & \texttt{google/gemma-2-2b} & 26 & 13 \\
SmolLM2-1.7B & \texttt{HuggingFaceTB/SmolLM2-1.7B} & 24 & 12 \\
Phi-3-mini & \texttt{microsoft/Phi-3-mini-4k-instruct} & 32 & 16 \\
OLMo-2-1B & \texttt{allenai/OLMo-2-0425-1B} & 16 & 8 \\
\bottomrule
\end{tabular}
\caption{Checkpoints, transformer-block counts, and analyzed layer indices.}
\label{tab:model_checkpoints}
\end{table}

All models use SDPA attention and tokenizers loaded from their corresponding checkpoint identifiers.

\section{Additional Evidence for Selective Ablation}
\label{app:selective}

\subsection{Cross-fitted ablation}
\label{app:crossfit}

We report balanced accuracy, the mean recall across the three timer classes, averaged over splits and models. Chance performance is $1/3$. In-sample removal of each timer's own plane yields balanced accuracies of $0.332$ for language and $0.334$ for topic, both approximately at chance. Table~\ref{tab:crossfit} reports the corresponding cross-fitted results for each model.

\begin{table}[!htbp]
\centering
\small
\begin{tabular}{lrrrrrr}
\toprule
 & \multicolumn{3}{c}{Language timer} & \multicolumn{3}{c}{Topic timer} \\
\cmidrule(lr){2-4}\cmidrule(lr){5-7}
Model & Intact & Own & Other & Intact & Own & Other \\
\midrule
Llama-3.2-1B     & 0.891 & 0.423 & 0.886 & 0.460 & 0.390 & 0.459 \\
Mistral-7B       & 0.917 & 0.433 & 0.909 & 0.513 & 0.403 & 0.506 \\
Qwen3-1.7B       & 0.878 & 0.421 & 0.867 & 0.502 & 0.404 & 0.498 \\
Qwen3-1.7B-Base  & 0.941 & 0.431 & 0.938 & 0.531 & 0.412 & 0.528 \\
Qwen3-4B         & 0.864 & 0.428 & 0.851 & 0.522 & 0.414 & 0.512 \\
pythia-1.4b      & 0.839 & 0.412 & 0.832 & 0.400 & 0.372 & 0.399 \\
gemma-2-2b       & 0.884 & 0.430 & 0.862 & 0.520 & 0.411 & 0.508 \\
SmolLM2-1.7B     & 0.963 & 0.437 & 0.962 & 0.452 & 0.391 & 0.452 \\
Phi-3-mini       & 0.857 & 0.416 & 0.835 & 0.511 & 0.404 & 0.505 \\
OLMo-2-1B        & 0.937 & 0.428 & 0.938 & 0.445 & 0.387 & 0.446 \\
\midrule
Mean             & 0.897 & 0.426 & 0.888 & 0.486 & 0.399 & 0.481 \\
\bottomrule
\end{tabular}
\caption{Cross-fitted balanced accuracy before ablation (Intact), after own-plane removal (Own), and after concurrent-timer-plane removal (Other).}
\label{tab:crossfit}
\end{table}

\subsection{Controls for document splitting}

Phase-matched splits assign the three document roles separately within each of the 12 combinations of language-switch and topic-change phase. Within-document splits use disjoint randomly assigned sentences from the same documents for plane estimation and decoder fitting.

\begin{table}[!htbp]
\centering
\small
\begin{tabular}{lrrrr}
\toprule
& Own, random & Own, phase-matched & Own, within-document & Other, random \\
\midrule
Language & 84.5\% & 84.0\% & 81.4\% & 1.6\% \\
Topic    & 58.4\% & 63.0\% & 53.6\% & 2.3\% \\
\bottomrule
\end{tabular}
\caption{Percentage of above-chance decoding signal removed under each split condition, averaged over per-model removal fractions from a separate eight-draw experiment. Own and Other denote removal of the target timer's plane and the concurrent timer's plane, respectively.}
\label{tab:split_controls}
\end{table}

\subsection{Past-event decoding without current-event states}

We decode timer values $1$ versus $2$ using the history-only procedure described above. Table~\ref{tab:history_summary} reports balanced accuracy averaged across splits and models, with chance at $0.5$, and the percentage of above-chance signal removed by each ablation.

\begin{table}[!htbp]
\centering
\small
\begin{tabular}{lrrrrr}
\toprule
Timer & Intact & Own, cross-fitted & Other, cross-fitted & Own removed & Other removed \\
\midrule
Language & 0.893 & 0.576 & 0.886 & 81\% & 1.7\% \\
Topic    & 0.598 & 0.553 & 0.591 & 45\% & 6.3\% \\
\bottomrule
\end{tabular}
\caption{Balanced accuracy and percentage of above-chance signal removed in history-only decoding. Own and Other denote ablation of the target timer's plane and the concurrent timer's plane, respectively.}
\label{tab:history_summary}
\end{table}

\subsection{Calibration with imposed overlap}

We transfer a fraction $\alpha$ of one timer's discriminative component into the concurrent timer's plane while keeping labels fixed. We measure the resulting decrease in balanced accuracy after removing the concurrent timer's plane (cross-ablation) or the target timer's own plane (self-ablation).

\begin{table}[!htbp]
\centering
\small
\begin{tabular}{rrrr}
\toprule
Transferred fraction $\alpha$ & Cross-ablation & Self-ablation & Above-chance accuracy \\
\midrule
0.00 & 0.6  & 16.5 & +0.164 \\
0.25 & 1.1  & 11.2 & +0.109 \\
0.50 & 4.4  & 8.1  & +0.081 \\
0.75 & 10.4 & 8.5  & +0.084 \\
1.00 & 11.0 & 10.8 & +0.110 \\
\bottomrule
\end{tabular}
\caption{Ablation effects under imposed overlap, reported as balanced-accuracy decreases in percentage points. The final column reports intact balanced accuracy minus chance.}
\label{tab:overlap_calibration}
\end{table}

\subsection{Aperiodic timing and analysis variants}

Aperiodic schedules independently sample inter-event gaps from $\{3,4,5\}$. The periodic control uses the same event markers; both conditions are subsampled to equalize the nine joint timer states. The mean ratio of cross-ablation to self-ablation, measured by their respective decreases in balanced accuracy, is $3.40\%$ for aperiodic schedules and $3.51\%$ for the matched periodic control.

Across alternative rank-2 plane estimators, cross-ablation is $1.12$--$5.08\%$ of self-ablation. With $64$--$150$ retained principal components, self-ablation is $29$--$88$ times larger than cross-ablation.

\section{Additional Evidence for Shared Information and Joint Geometry}
\label{app:joint}

\subsection{Restricted decoding and principal angles}

Restricted decoding trains and evaluates a decoder using representations projected onto a rank-2 plane. Table~\ref{tab:restricted_per_model} reports language-timer balanced accuracy minus chance ($1/3$) and the smaller principal angle between the language and topic planes.

\begin{table}[!htbp]
\centering
\small
\begin{tabular}{lrrrr}
\toprule
Model & Own plane & Concurrent plane & Placebo & Angle \\
\midrule
SmolLM2-1.7B     & +0.596 & +0.136 & +0.164 & $66.6^\circ$ \\
Qwen3-1.7B-Base  & +0.537 & +0.259 & +0.149 & $53.4^\circ$ \\
OLMo-2-1B        & +0.528 & +0.188 & +0.146 & $52.4^\circ$ \\
Mistral-7B-v0.3  & +0.515 & +0.206 & +0.166 & $53.1^\circ$ \\
Llama-3.2-1B     & +0.451 & +0.153 & +0.117 & $55.4^\circ$ \\
gemma-2-2b       & +0.438 & +0.217 & +0.105 & $41.6^\circ$ \\
Qwen3-1.7B       & +0.379 & +0.126 & +0.071 & $56.3^\circ$ \\
Qwen3-4B         & +0.360 & +0.134 & +0.078 & $51.9^\circ$ \\
Phi-3-mini       & +0.326 & +0.144 & +0.061 & $46.2^\circ$ \\
pythia-1.4b      & +0.270 & +0.095 & +0.053 & $62.4^\circ$ \\
\bottomrule
\end{tabular}
\caption{Above-chance language-timer decoding from the language plane (Own), topic plane (Concurrent), and placebo plane, with the smaller principal angle between the two timer planes.}
\label{tab:restricted_per_model}
\end{table}

The smaller principal angle averages $53.9^\circ$ across models, compared with $79.7^\circ$ for unrelated rank-2 planes.

\subsection{Whitening and the shared direction}

Whitening within-class scatter before plane estimation increases the timer-plane angle to the range observed for unrelated subspaces and reduces cross-timer restricted decoding.

The most closely aligned direction carries $5.1$ times the variance of a random direction. Language-timer balanced accuracy along this direction is $0.206$ above chance, approximately $47\%$ of the above-chance accuracy in the full language plane. Removing this direction increases the angle between the remaining plane directions from $54.8^\circ$ to $78.3^\circ$.

\subsection{Joint additivity and cross-document prediction}

Table~\ref{tab:additivity} reports the interaction residual defined in Section~\ref{sec:measures}, its finite-sample additive baseline, and held-out variance explained under the cross-document prediction procedure.

\begin{table}[!htbp]
\centering
\small
\begin{tabular}{lrrrr}
\toprule
 & \multicolumn{2}{c}{Interaction residual} & \multicolumn{2}{c}{Held-out variance explained} \\
\cmidrule(lr){2-3}\cmidrule(lr){4-5}
Model & Observed & Additive floor & Additive & Saturated \\
\midrule
Llama-3.2-1B     & 9.1\% & 4.3\% & 0.839 & 0.833 \\
Mistral-7B       & 6.8\% & 3.4\% & 0.876 & 0.852 \\
Qwen3-1.7B       & 12.8\% & 6.0\% & 0.758 & 0.761 \\
Qwen3-1.7B-Base  & 7.5\% & 3.1\% & 0.868 & 0.874 \\
Qwen3-4B         & 12.8\% & 6.1\% & 0.767 & 0.753 \\
pythia-1.4b      & 16.8\% & 11.2\% & 0.664 & 0.511 \\
gemma-2-2b       & 10.4\% & 4.4\% & 0.813 & 0.816 \\
SmolLM2-1.7B     & 5.2\% & 2.9\% & 0.897 & 0.865 \\
Phi-3-mini       & 13.3\% & 7.3\% & 0.749 & 0.717 \\
OLMo-2-1B        & 6.6\% & 3.5\% & 0.871 & 0.838 \\
\midrule
Mean             & 10.1\% & 5.2\% & 0.810 & 0.782 \\
\bottomrule
\end{tabular}
\caption{Joint-centroid interaction residuals and held-out variance explained by additive and saturated nine-centroid models. The additive floor estimates the finite-sample residual under exactly additive population means.}
\label{tab:additivity}
\end{table}

Shuffling timer labels raises the interaction residual to approximately $50\%$. The mean-only model yields a mean held-out explained variance of $-0.005$.

\subsection{Conditional stability of timer displacements}

For each timer transition, we compute joint-centroid displacements at each value of the concurrent timer and measure their cosine similarity. We also measure split-half reproducibility of the same transition at a fixed concurrent-timer value and repeat the analysis with shuffled timer labels.

\begin{table}[!htbp]
\centering
\small
\begin{tabular}{lrrrr}
\toprule
Timer & Across & Split-half & Across / split-half & Shuffled \\
\midrule
Language & 0.855 & 0.889 & 96\% & +0.005 \\
Topic    & 0.331 & 0.587 & 54\% & +0.005 \\
\bottomrule
\end{tabular}
\caption{Mean cosine similarity of timer-transition displacements across concurrent-timer values (Across), between document halves at a fixed concurrent-timer value (Split-half), and under shuffled labels. Across / split-half reports the mean of per-model ratios.}
\label{tab:conditional_summary}
\end{table}
\section{Additional Generalization and Behavioral Results}
\label{app:extensions}

\subsection{Human-written dialogue corpus}

We analyze 3,000 DailyDialog conversations \citep{li2017dailydialog} with two timers: utterances since the most recent occurrence of the most common dialogue act, and utterances since the most recent utterance labeled with one of the six named non-neutral emotions. Event definitions are selected before extracting model states.

Across Llama-3.2-1B, Qwen3-1.7B, and pythia-1.4b, both timers are decodable at $28.0$--$31.4$ balanced-accuracy points above chance. Mean balanced-accuracy decreases are $30.1$ points for self-ablation and $3.79$ points for cross-ablation, giving a cross/self ratio of $12.6\%$.

\subsection{More event streams and shared event-recency geometry}

We analyze six event realizations---language, speaker tag, topic tag, punctuation, register phrase, and symbol prefix---in six single-timer corpora across four models. Within each split, the six training corpora share a fitted 100-component PCA basis. We estimate a pooled rank-2 timer plane across event realizations and a separate rank-2 plane for each realization (Figure~\ref{fig:shared_geometry}).

\begin{figure}[!htbp]
\floatconts
  {fig:shared_geometry}
  {\caption{\textbf{Shared and event-specific event-recency geometry.}
  \textbf{A}, Restricted decoding, including event-specific residual planes formed by removing the pooled component.
  \textbf{B}, Principal angles between pooled and event-specific planes.
  \textbf{C}, Decoding after plane removal.
  Pale intervals show variation across four models.}}
  {\includegraphics[width=0.92\textwidth]{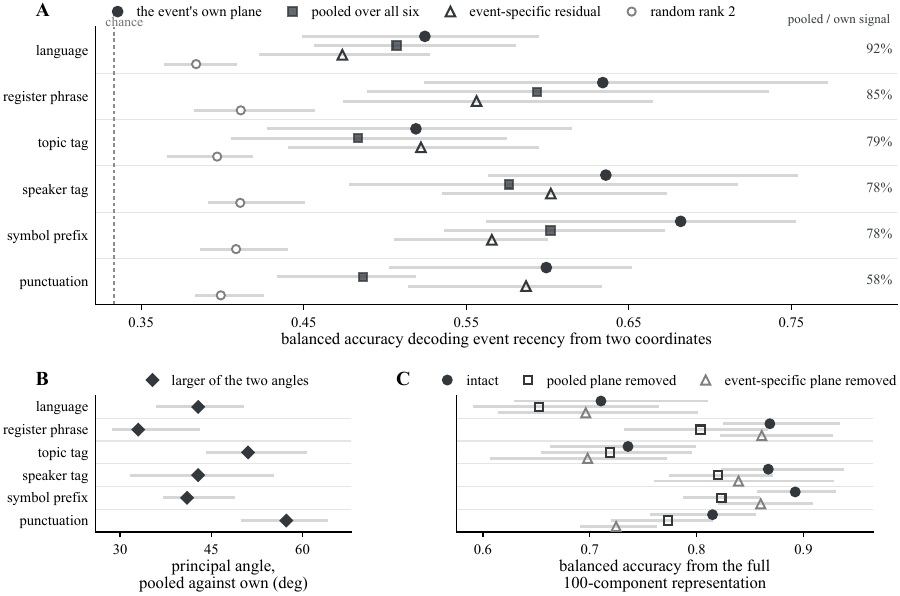}}
\end{figure}

\subsection{Longer timer horizons}
\label{app:horizons}

For horizon $H$, language switches have period $H+1$ and topic changes have period $H+2$. Excluding the final topic phase gives both timers analyzed values $0,\ldots,H$. We use $H\in\{2,5,10,20,50\}$ and 200 documents per horizon, with 24 propositions per topic in parallel English and German realizations. Empirical association between timer labels remains below $0.06$ in magnitude at every horizon.

Ridge regressors predict normalized timer value. On the plane-estimation fold, we decompose the centered timer centroids by SVD and select the smallest rank explaining $90\%$ of their variance. Under three-way cross-fitting, the selected subspace is removed before decoder training and evaluation on separate folds. Self-, cross-, and isotropic-random ablations use the same selected rank.

Table~\ref{tab:horizon_crossfit} reports the horizon sweep.
For each model, removal is calculated as
$100(\rho_{\mathrm{intact}}-\rho_{\mathrm{ablated}})/
\rho_{\mathrm{intact}}$ using split-averaged correlations,
then averaged across models. Cross/self is the mean of
per-model removal ratios.

Split-half subspace angles range from $10.6^\circ$ to $14.5^\circ$. Shuffled-label reference angles decrease from $78.4^\circ$ to $50.5^\circ$ as selected rank increases.

\subsection{Within-family parameter range}

We analyze Qwen3 models with $0.6$B, $1.7$B, $4$B, $8$B, and $14$B parameters, holding the corpus, layer-selection rule, preprocessing, projection dimension, splits, and decoder fixed. All five models use 4-bit weights; the three smaller models are also evaluated in bfloat16. Interaction residuals use the full-corpus estimator, and split-half plane angles are averaged over eight splits.

\begin{table}[!htbp]
\centering
\small
\begin{tabular}{lrrr}
\toprule
Model & Interaction $\pm$ SEM & Shuffled & Split-half angle (lang./topic) \\
\midrule
Qwen3-0.6B  & $12.1 \pm 1.7\%$ & 50.4\% & $17.2^\circ/58.7^\circ$ \\
Qwen3-1.7B  & $14.6 \pm 2.1\%$ & 50.6\% & $22.2^\circ/51.7^\circ$ \\
Qwen3-4B    & $15.9 \pm 2.4\%$ & 49.9\% & $27.0^\circ/51.7^\circ$ \\
Qwen3-8B    & $13.3 \pm 2.1\%$ & 50.1\% & $19.7^\circ/53.4^\circ$ \\
Qwen3-14B   & $12.7 \pm 2.0\%$ & 50.3\% & $20.2^\circ/41.1^\circ$ \\
\bottomrule
\end{tabular}
\caption{Interaction residuals and plane reproducibility across Qwen3 model sizes. Uncertainty is one bootstrap standard error estimated from 40 document resamples; Shuffled reports the shuffled-label interaction residual. Shuffled-label plane angles range from $84^\circ$ to $86^\circ$.}
\label{tab:qwen_ladder}
\end{table}

At the three overlapping sizes, 4-bit and bfloat16 interaction estimates differ by at most $0.9$ percentage points and language-plane angles by at most $1.2^\circ$.

\subsection{Behavioral continuation assay and output intervention}

For context $x$, let $s(x)$ be the summed log-probability difference between other-language and current-language next-sentence candidates expressing the same proposition and topic. We compare language-switch periods $k=3$ and $k=4$ at matched timer value $e$:
\[
A_{\mathrm{due}}=s(k=3,e=2)-s(k=4,e=2),\qquad
A_{\mathrm{control}}=s(k=3,e=1)-s(k=4,e=1).
\]
We report $D=A_{\mathrm{due}}-A_{\mathrm{control}}$ over 140 documents per schedule for each of the ten pretrained checkpoints.

\begin{table}[!htbp]
\centering
\scriptsize
\begin{tabular}{lrrr}
\toprule
Model & $A_{\mathrm{due}}$ & $A_{\mathrm{control}}$ & $D \pm$ SEM \\
\midrule
pythia-1.4b            & 3.301 & -0.837 & $4.138 \pm 0.658$ \\
Qwen3-1.7B             & 2.147 & -0.741 & $2.888 \pm 0.556$ \\
Phi-3-mini             & 2.094 & -0.750 & $2.844 \pm 0.456$ \\
OLMo-2-1B              & 1.997 & -0.767 & $2.764 \pm 0.569$ \\
Llama-3.2-1B           & 2.262 & -0.280 & $2.542 \pm 0.418$ \\
SmolLM2-1.7B           & 1.962 & -0.449 & $2.411 \pm 0.546$ \\
Qwen3-4B               & 2.256 &  0.006 & $2.250 \pm 0.497$ \\
gemma-2-2b             & 1.224 & -0.754 & $1.978 \pm 0.487$ \\
Qwen3-1.7B-Base        & 1.572 & -0.355 & $1.927 \pm 0.358$ \\
Mistral-7B-v0.3        & 1.567 & -0.210 & $1.776 \pm 0.354$ \\
Qwen3-1.7B-random      & -2.427 &  1.362 & $-3.789 \pm 0.863$ \\
\bottomrule
\end{tabular}
\caption{Recency-matched continuation contrasts. Uncertainty is one standard error from 2000 document-bootstrap resamples. The final row reports a randomly initialized Qwen3-1.7B control.}
\label{tab:behavior_gate}
\end{table}

We estimate the language-timer displacement from value 1 to value 2 on 70 documents and apply it at 490 held-out sentence boundaries per pretrained checkpoint. Interventions modify the sentence-final token's middle-layer state at strengths $0.5$ and $1.0$; concurrent-timer and isotropic controls are norm-matched. Table~\ref{tab:behavior_steer} reports full-strength changes in other-language versus current-language continuation log odds.

\begin{table}[!htbp]
\centering
\scriptsize
\begin{tabular}{lrrrr}
\toprule
Model & Target & Concurrent & Random & Target $-$ concurrent \\
\midrule
Phi-3-mini       &  0.0249 &  0.0231 &  0.0010 &  0.0018 \\
gemma-2-2b       &  0.0243 &  0.0066 & -0.0070 &  0.0177 \\
OLMo-2-1B        &  0.0173 &  0.0499 & -0.0015 & -0.0326 \\
pythia-1.4b      &  0.0126 & -0.0164 &  0.0037 &  0.0290 \\
Mistral-7B-v0.3  &  0.0046 &  0.0215 & -0.0036 & -0.0169 \\
SmolLM2-1.7B     &  0.0021 &  0.0324 & -0.1014 & -0.0303 \\
Qwen3-4B         & -0.0017 & -0.0125 & -0.0018 &  0.0108 \\
Llama-3.2-1B     & -0.0105 &  0.0166 & -0.0150 & -0.0271 \\
Qwen3-1.7B-Base  & -0.0113 & -0.0182 & -0.0163 &  0.0069 \\
Qwen3-1.7B       & -0.0142 & -0.0087 & -0.0077 & -0.0055 \\
\midrule
Mean             &  0.0048 &  0.0094 & -0.0150 & -0.0046 \\
\bottomrule
\end{tabular}
\caption{Mean continuation log-odds change at strength $1.0$ for language-timer (Target), concurrent-timer, and isotropic (Random) interventions.}
\label{tab:behavior_steer}
\end{table}

\end{document}